\documentclass{article}

\usepackage[table]{xcolor}
\usepackage{multirow}

\usepackage{rotating}
\usepackage{graphicx}
\usepackage{lscape}

\usepackage{PRIMEarxiv}
\usepackage{amsmath}
\usepackage{authblk}
\usepackage[utf8]{inputenc} 
\usepackage[T1]{fontenc}    
\usepackage{hyperref}       
\usepackage{url}            
\usepackage{booktabs}       
\usepackage{amsfonts}       
\usepackage{nicefrac}       
\usepackage{microtype}      
\usepackage{lipsum}
\usepackage{fancyhdr}       
\usepackage{graphicx}       
\graphicspath{{media/}}     

\title{Leveraging Large Language Models for Enhanced NLP Task Performance through Knowledge Distillation and Optimized Training Strategies
\thanks{\textit{\underline{Corresponding author}
}: 
\textbf{Yining Huang \  email: \ huangyining1987@gmail.com}} 
}

\author[\space\space1 ]{Yining Huang \thanks{huangyining1987@gmail.com}}
\author[\space\space2,3]{Keke Tang  \thanks{tkk2012@gmail.com}}
\author[\space\space]{Meilian Chen \thanks{523062863@qq.com}}
\affil[1]{School of Politics and Public Administration, South China Normal University}
\affil[2]{University of Chinese Academy of Sciences}
\affil[3]{Shenyang institute of computing technology, Chinese academy of sciences}

\begin{document}
\maketitle

\begin{abstract}
Emerging Large Language Models (LLMs) like GPT-4 have revolutionized Natural Language Processing (NLP), showing potential in traditional tasks such as Named Entity Recognition (NER). Our study explores a three-phase training strategy that harnesses GPT-4's capabilities to enhance the BERT model's performance on NER. Initially, GPT-4 annotates a subset of the CONLL2003 and additional BBC dataset without fine-tuning. We then train BERT using a mix of original and LLM-annotated data, analyzing the efficacy of LLM annotations against traditional methods.

The second phase involves comparative experiments with different training regimens, assessing the synergy between distilled and original data. We observe that sequential strategies, particularly a simple mix of training first with distilled data followed by original data, significantly boost performance. In the third phase, we investigate various data blending techniques, including sigmoid and power decay functions, to optimize the training process further. Our results indicate that a strategic mix of distilled and original data markedly elevates the NER capabilities of BERT.

Our approach presents a scalable methodology that reduces manual annotation costs and increases efficiency, making it especially pertinent in resource-limited and closed-network environments. The study concludes that while the 'Simple Mix' strategy yields the best results, understanding its underlying mechanisms requires further research. Future work will also focus on refining prompt designs and enhancing annotation selection processes, aiming to extend our methodology to diverse NLP tasks.
\end{abstract}


\section*{Introduction}
The advent of Large Language Models (LLMs) such as the Generative Pre-trained Transformer (GPT) series has marked a significant milestone in the field of artificial intelligence and natural language processing (NLP). With their remarkable reasoning capabilities and extensive knowledge bases, LLMs have demonstrated unparalleled performance across a wide range of applications, from text generation and dialogue systems to content comprehension. Despite their success, these models face a notable performance gap in traditional NLP tasks like Named Entity Recognition (NER) and Relation Extraction (RE) when compared to the current state-of-the-art (SOTA) technologies. However, with continuous advancements, there is a promising potential for LLMs to not only bridge this gap but also to significantly reduce costs and enhance efficiency in traditional NLP tasks.

The rising dominance of LLMs has sparked debates regarding the relevance of traditional NLP techniques. Some argue that the capabilities of LLMs might render traditional methods obsolete. Yet, this perspective overlooks the intrinsic limitations of LLMs, such as their capacity constraints and tendency towards hallucination—producing outputs that may diverge from factual accuracy. In reality, traditional NLP techniques remain indispensable, especially in applications such as knowledge graph construction where tasks like NER and RE are crucial for extracting structured information from unstructured text. Therefore, LLMs and traditional NLP techniques are not in competition but rather complement each other, forming a synergistic relationship that leverages the strengths of both approaches.

However, integrating LLMs into traditional NLP tasks is not without challenges. First, the inherent differences in task nature between generative models like LLMs and the sequence labeling or classification tasks of traditional NLP present a significant gap. Bridging this gap requires innovative approaches that have yet to be fully explored. Additionally, LLMs are prone to producing "hallucinations," which can compromise the accuracy of generated annotations. This issue is exacerbated by the dependency of output quality on the design of prompts, introducing variability and potential inaccuracies in task execution. Moreover, the substantial computational resources required for LLMs pose a barrier to their application in resource-constrained environments or scenarios with stringent privacy and security requirements, necessitating solutions that can operate within these limitations.

Addressing these challenges, our research proposes a comprehensive solution utilizing knowledge distillation techniques. By leveraging LLMs for data annotation and incorporating these annotations in the training of smaller models like BERT, we enhance the performance of these models on traditional NLP tasks. Our approach mitigates the limitations of LLMs through the use of Chain of Thought (CoT) prompting to refine reasoning processes and generate more accurate annotations. Furthermore, the output format in the prompt for instructing the LLM is more efficient for generative model to predict nested or discontinues entities than merely outputting sequence of tags. 

Through a series of meticulously designed experiments, we have undertaken a comprehensive evaluation of BERT model training strategies, leveraging both GPT-4 annotated (distilled) data and traditional human-annotated data. Our multifaceted approach included exploring the impact of various training regimens: training exclusively with distilled data, exclusively with original data, and a combination of both in varying sequences and proportions. We investigated the potential of distilled data from similar domains to enhance performance and employed innovative data blending techniques to optimize training efficacy. The results from these experiments have been illuminating, consistently demonstrating that strategic training with a mix of distilled and original data can significantly elevate the model's NER capabilities. Notably, our findings suggest that simple sequential mixing of data types and fine-tuning the proportion of data sources throughout the training process can lead to substantial performance gains. 

This framework not only validates the feasibility of enhancing traditional NLP tasks with LLM-generated annotations but also showcases a scalable methodology that can be adapted across various LLMs, smaller models, and datasets. Our findings underscore the potential of this approach to reduce manual annotation costs, increase efficiency, and enable the deployment of capable models in resource-limited and privacy-sensitive settings, thereby filling a critical gap left by the limitations of LLMs in these contexts.

\section*{Related works}
\label{sec:headings}

\paragraph{Large Language Model} The integration of Large Language Models (LLMs) \cite{bert,alm,martin2019camembert,ganlm,gtrans,claude,llama,llama2,gpt4,chai2024xcot,wang2023mac,yin2023finpt,chai2023qurg,low_resource_template,lvpm3,chai2023qurg} into Named Entity Recognition (NER) tasks have seen varied approaches and methodologies, demonstrating the evolving landscape of NLP research in leveraging advanced AI models for enhancing traditional tasks. This section outlines related work that has contributed to the development of methods and strategies in applying LLMs and other advanced techniques to NER, highlighting the diverse efforts to address challenges such as data scarcity, model efficiency, and cross-lingual applications.
Many LLMs further improve the multilingual natural language processing and understanding tasks by supervised fine-tuning \cite{bai2023knowprefix,wang2023multilingual,hlt_mt,chai2024xcot,chen2024orion}.

\paragraph{Named Entity Recognition} Cross-lingual NER remains a significant challenge due to the scarcity of annotated training data, especially in low-resource languages. To minimize the gap among different languages, some researchers \cite{jia2020entity,wmt2021,al2015polyglot，yang2022crop,wang2023mt4crossoie,ho2022large,wei2022emergent}, introduce the translation-based framework to enhance zero-shot cross-lingual NER, demonstrating substantial improvements over previous approaches. The gap between the sequence labeling nature of NER tasks and the text-generation orientation of LLMs is explored in \cite{wang2304gpt,yang2021learning,rahimi2019massively,beltagy2019scibert,mo2024c}, where the GPT-NER model is proposed. By transforming the NER task into a text generation format, GPT-NER addresses the hallucination issue common in LLMs, showing promise in low-resource and few-shot settings. PromptNER \cite{ashok2023promptner} presents a novel few-shot and cross-domain NER algorithm that utilizes prompt-based heuristics with LLMs, requiring only a few-shot examples and entity type definitions to achieve state-of-the-art performance on several benchmarks. This underscores the potential of prompt-based methods in NER tasks. \cite{zhou2023universalner,vsuppa2021benchmarking} explores targeted distillation with mission-focused instruction tuning to train student models for NER tasks, demonstrating that these models can significantly outperform general instruction-tuned models across a broad application class, including open information extraction. The Multi-view Contrastive Learning approach for Cross-lingual NER \cite{liu2018empower,mo2023mcl,chou2020construction,yang2021learning} tackles the challenge of semantic and token-level representation alignment across languages, achieving new state-of-the-art performance on the XTREME benchmark for 40 languages. Inspired by the Chain of Thought methodology, \cite{bian2023inspire} leverages the step-by-step problem-solving capabilities of LLMs for Biomedical NER, incorporating external knowledge to enhance entity type determination, showcasing significant improvements in a domain-specific application. \cite{xie2023empirical} focuses on zero-shot information extraction in NER, adapting reasoning strategies tailored for NER tasks with LLMs. The introduction of syntactic augmentation and a two-stage majority voting strategy demonstrates remarkable improvements across multiple benchmarks. Lastly, \cite{amalvy2023learning} addresses the issue of applying pre-trained transformer models to long documents for NER tasks. By generating a synthetic context retrieval training dataset and training a neural context retriever based on a BERT model, this work shows that it is possible to outperform unsupervised retrieval baselines, marking an advancement in handling long-document NER tasks. These works collectively underscore the dynamic interplay between LLMs, traditional NLP techniques, and innovative methodologies in advancing the field of NER. Our research builds upon these foundations, aiming to further leverage the capabilities of LLMs to enhance NER model training and performance, particularly through the integration of Chain of Thought prompting and knowledge distillation techniques.

\section*{Methodology}
\label{sec:headings}

\subsection*{Overview}
This research delineates a three-phase exploration into augmenting Named Entity Recognition (NER) performance by integrating Large Language Models (LLMs), specifically GPT-4, for data annotation, and subsequently training a BERT model with these annotations. Initially, we assessed GPT-4's annotation capabilities on a subset of the CONLL2003 dataset\cite{tjong-kim-sang-de-meulder-2003-introduction}, employing both standard and Chain of Thought (CoT) prompting techniques to measure the quality of LLM-generated annotations against human annotations. Following this, we conduct a comparative training experiment to evaluate the impact of LLM annotations on BERT's NER capabilities, using a blend of original and LLM-annotated data. The study further innovates by incorporating external, similar domain data (from the BBC news dataset) to enrich the training material, thus exploring the potential of external datasets to enhance model performance. The third part of our experiment involves a examination of various training strategies to determine the most effective methodology for integrating distilled and original data. This approach aims to provide insights into the benefits and limitations of leveraging LLM annotations within traditional NLP frameworks. The data and code supporting the findings of this study have been made publicly available and can be accessed online\footnote{\url{https://huggingface.co/datasets/kobe1987/DLLM2TM}}.

\subsubsection*{Benchmark Dataset}
Our study employs the CONLL2003 dataset, a cornerstone in the field of named entity recognition, focusing on four types of named entities: persons, locations, organizations, and miscellaneous entities that do not fit into the aforementioned categories.  

For the initial phase of our experiment, we selected a random sample of 1000 sentences from the training set of the CONLL2003 dataset. This subset serves as the basis for assessing the annotation capabilities of LLM, aiming to explore the model's proficiency in generating annotations comparable to the dataset's meticulously crafted tags. Following the annotation process, the distilled data, enriched with LLM's insights, along with another randomly selected sample of 1000 sentences from the original training set, are utilized in a mixed training approach to fine-tune a smaller model. The effectiveness of this strategy is subsequently evaluated on the CONLL2003 test set, allowing us to measure the impact of leveraging LLM annotations on model performance in named entity recognition tasks.

\subsection*{Phase One: Data Annotation with LLM}

\subsubsection*{LLM Annotation}
We employed GPT-4, specifically the gpt-4-0125-preview model provided by OpenAI, as our chosen Large Language Model (LLM) for annotating a subset of 1000 randomly sampled sentences from the CONLL2003 dataset. GPT-4 represents a significant leap forward in the domain of AI-driven natural language understanding and generation. As a large multimodal model, it has demonstrated unparalleled proficiency in accepting both text and image inputs, outputting text that showcases broader general knowledge and more advanced reasoning capabilities than its predecessors.

For our annotation process, we leveraged the GPT-4 turbo API to meticulously generate named entity annotations for the selected CONLL2003 data samples. This process involved employing two distinct prompting methods (detailed in subsequent sections and Figure~\ref{fig:standard_Prompting_Templates} and Figure~\ref{fig:cot_Prompting_Templates}), which facilitated the model's understanding and execution of the annotation task. While GPT-4's outputs were predominantly accurate and aligned with the task requirements, a minimal amount of manual post-processing was necessary to rectify occasional irregularities in the generated annotations, ensuring the highest quality of data for our study.

\subsubsection*{Standard Prompting}
We employed a few-shot prompting approach, shown in Figure~\ref{fig:standard_Prompting_Templates}, with GPT-4 to annotate a random sample of 1000 sentences from the CONLL2003 dataset, serving as a baseline for comparison. Few-shot learning, a method where the model learns to perform tasks from a limited number of examples, was utilized to guide GPT-4 in accurately identifying and classifying named entities into four categories: LOC (Location), MISC (Miscellaneous), ORG (Organization), and PER (Person).

The prompt designed for this task succinctly introduced GPT-4 to its role as an NLP annotation expert, tasked with determining the type of each named entity present in a given sentence. It outlined the definitions for the four entity types, emphasizing the importance of showing the reasoning process in steps before presenting the results in a structured format. This approach aimed to leverage GPT-4's advanced reasoning capabilities and vast general knowledge for generating precise entity annotations.

Included in the prompt were several example sentences, each followed by their expected annotation results, showcasing how entities should be categorized according to the predefined types. This structured prompt not only directed GPT-4 on how to approach the annotation task but also provided clear examples of the task's expected outcomes, facilitating a more accurate and efficient annotation process.

\subsubsection*{CoT Prompting}
\begin{figure}[hbt!]
\centering
\includegraphics[width=0.95\linewidth]{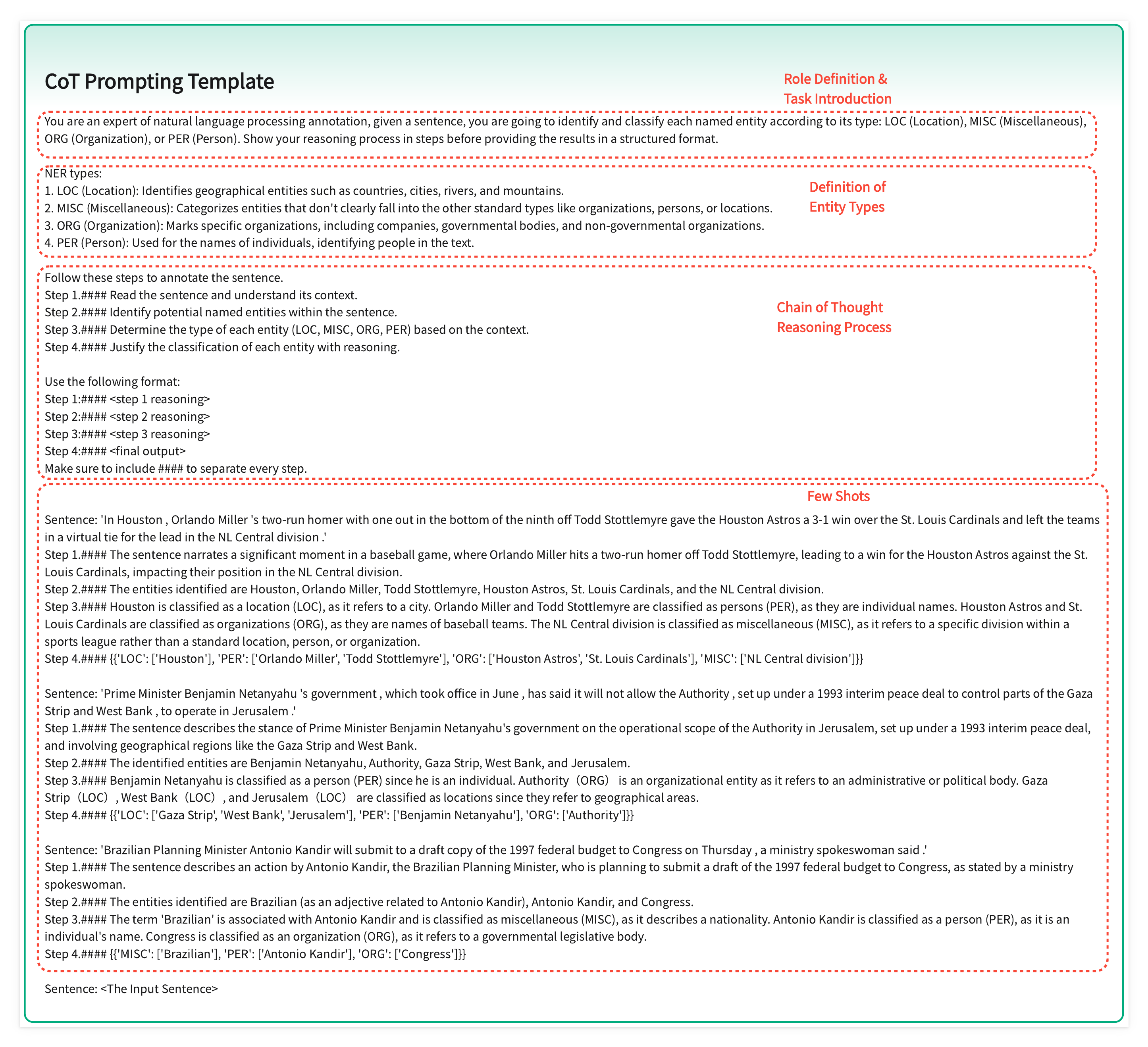}
\caption{The CoT prompting Template}
\label{fig:cot_Prompting_Templates}
\end{figure}

Alongside the standard prompting approach, we implemented the Chain of Thought (CoT) prompting method combined with few-shot learning to annotate the same 1000 randomly sampled sentences from the CONLL2003 dataset using GPT-4, which is shown in Figure~\ref{fig:cot_Prompting_Templates}, . The CoT methodology is designed to mimic human-like reasoning by guiding the model through a stepwise thought process before arriving at a conclusion. This approach not only aims to improve the accuracy of the annotations by making the model's thought process explicit but also seeks to provide insights into the model's reasoning, thereby enhancing the interpretability of the annotations.

The CoT prompt we developed instructs GPT-4 to follow a four-step reasoning process for each sentence. It begins with understanding the context of the sentence, identifying potential named entities, determining the type of each entity based on context, and justifying the classification with reasoning. This structured approach encourages GPT-4 to articulate its thought process at each step, culminating in the structured format of named entity recognition results.

For instance, in annotating a sentence about a baseball game, GPT-4 is guided to recognize and classify entities such as "Houston" as a location (LOC) and "Orlando Miller" and "Todd Stottlemyre" as persons (PER), with justification provided for each classification. This not only aligns with the task's requirements but also allows for an examination of how GPT-4 applies general knowledge and contextual reasoning to solve specific NLP tasks.

The use of CoT in conjunction with few-shot examples serves to further anchor GPT-4's responses within the task's framework, leveraging both the model's advanced language understanding capabilities and its ability to engage in complex reasoning. By comparing the annotations generated through standard prompting with those produced via the CoT method, we aim to assess the impact of explicit reasoning on the quality and reliability of LLM-generated annotations for named entity recognition tasks.

\begin{figure}[hbt!]
\centering
\includegraphics[width=0.95\linewidth]{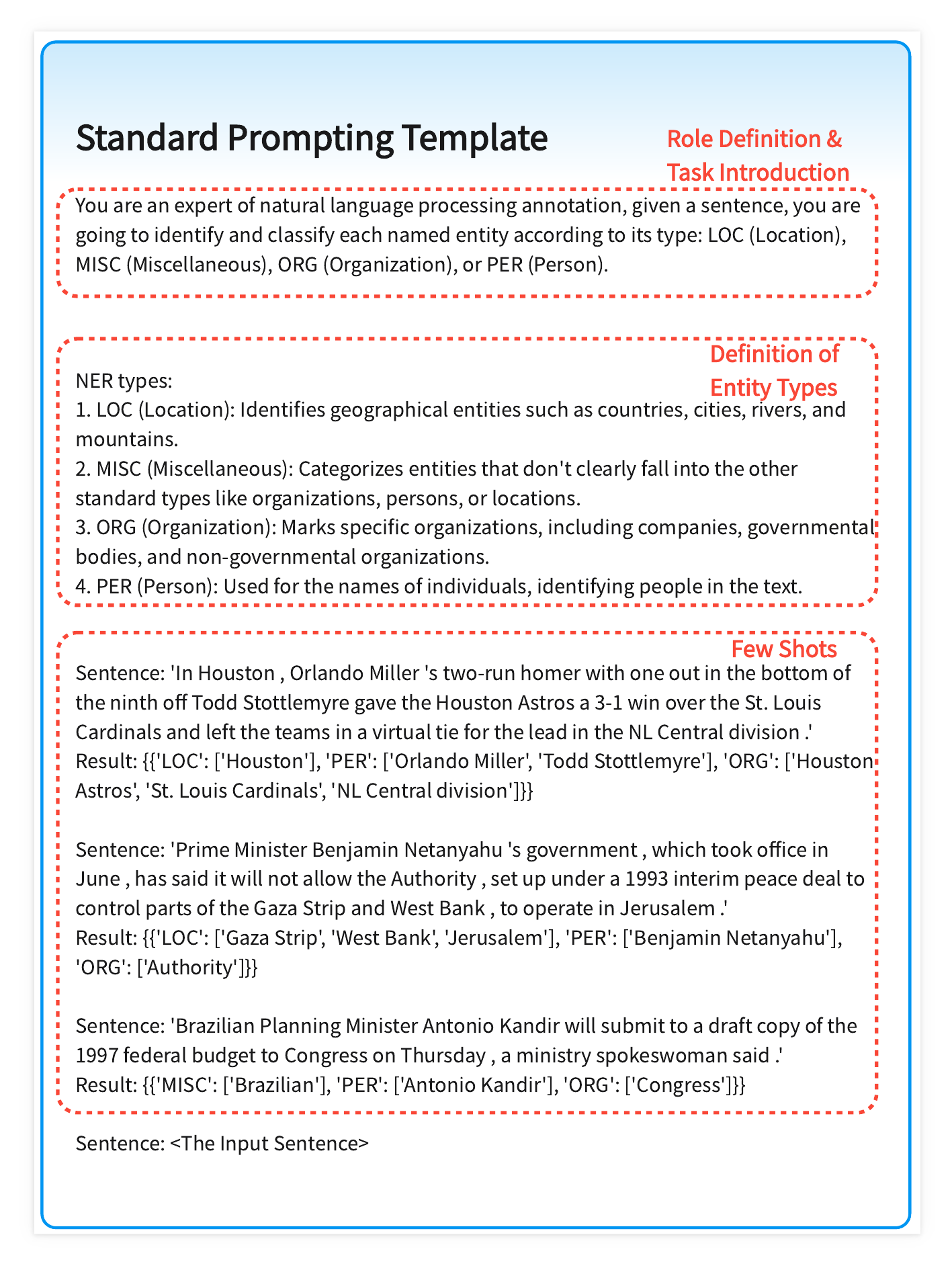}
\caption{The Standard prompting Template}
\label{fig:standard_Prompting_Templates}
\end{figure}

Plus, as shown in both standard prompting and CoT prompting in Figure~\ref{fig:standard_Prompting_Templates} and Figure~\ref{fig:cot_Prompting_Templates}, by utilizing a structured dictionary output format, LLMs can effectively capture and denote complex entity relationships within text. For instance, in the sentence "my fingers swelled up and hurt", LLMs can identify the discontinuous entity by outputting \{'symptom': 'fingers swelled up', 'fingers hurt'\}, encapsulating the related but non-adjacent terms. Similarly, for nested entities such as in "national basketball of America"(a made up organization name), the model can differentiate and simultaneously output the encompassing organization and the included location with \{'ORG': 'national basketball of America', 'LOC': 'America'\}, demonstrating the prompt's ability to recognize and represent nested entities accurately.


\subsubsection*{Results and Analysis}
Upon employing two distinct prompting strategies for GPT-4 annotation on a subset of 1000 sentences from the CONLL2003 dataset, our analysis yielded insightful results. The implementation of the Chain of Thought (CoT) method, which promotes an articulated step-by-step reasoning process, has demonstrated a pronounced advantage in performance over the standard few-shot approach.
The results, as depicted in Table~\ref{tab:performance_comparison}, indicate that the CoT prompting method achieved an overall F1-score of 0.73, which is superior to the 0.65 F1-score obtained using the standard few-shot prompting method. More specifically, the CoT method showed considerable improvements in precision and recall across most named entity types, with the most notable enhancement seen in the MISC category, where the CoT method's precision more than doubled that of the few-shot method.
This comparative performance analysis underscores the effectiveness of the CoT prompting method in mitigating the hallucination issue typically associated with LLMs. By prompting GPT-4 to engage in a transparent reasoning process, CoT prompting helped to align the model's outputs more closely with the expected annotations, thereby improving the quality and reliability of the annotations for NER tasks.
The higher performance metrics associated with CoT prompting suggest that guiding LLMs through a structured reasoning pathway can significantly amplify their reasoning abilities in complex NER tasks. This finding is pivotal, as it not only reinforces the value of CoT in enhancing the interpretability of LLM outputs but also demonstrates the potential for more accurate NER annotations, which is critical for downstream applications that rely on precise entity recognition.

\begin{table}[h!]
\centering
\begin{tabular}{|c|c|c|c|}
\hline
\rowcolor[HTML]{EFEFEF} 
\multicolumn{2}{|c|}{\textbf{Metric}} & \textbf{Standard Prompting} & \textbf{CoT Prompting} \\ \hline
\multirow{5}{*}{Overall} & Precision & 0.56 & \textcolor{red}{0.67} \\ \cline{2-4} 
                         & Recall & 0.78 & \textcolor{red}{0.82} \\ \cline{2-4} 
                         & Micro Average F1 & 0.65 & \textcolor{red}{0.73} \\ \cline{2-4} 
                         & Macro Average F1 & 0.67 & \textcolor{red}{0.72} \\ \cline{2-4} 
                         & Weighted Average F1 & 0.73 & \textcolor{red}{0.77} \\ \hline
\multirow{3}{*}{LOC}     & Precision & 0.73 & \textcolor{red}{0.81} \\ \cline{2-4} 
                         & Recall & \textcolor{red}{0.89} & 0.88 \\ \cline{2-4} 
                         & F1-Score & 0.80 & \textcolor{red}{0.84} \\ \hline
\multirow{3}{*}{ORG}     & Precision & 0.73 & \textcolor{red}{0.77} \\ \cline{2-4} 
                         & Recall & 0.67 & \textcolor{red}{0.76} \\ \cline{2-4} 
                         & F1-Score & 0.70 & \textcolor{red}{0.77} \\ \hline
\multirow{3}{*}{PER}     & Precision & \textcolor{red}{0.97} & 0.95 \\ \cline{2-4} 
                         & Recall & 0.96 & 0.96 \\ \cline{2-4} 
                         & F1-Score & \textcolor{red}{0.97} & 0.95 \\ \hline
\multirow{3}{*}{MISC}    & Precision & 0.13 & \textcolor{red}{0.23} \\ \cline{2-4} 
                         & Recall & 0.43 & \textcolor{red}{0.52} \\ \cline{2-4} 
                         & F1-Score & 0.20 & \textcolor{red}{0.31} \\ \hline
\end{tabular}
\caption{Performance comparison between Standard and CoT Prompting}
\label{tab:performance_comparison}
\end{table}

\subsection*{Phase Two: Training with Distilled Data}
\subsubsection*{Model Introduction}
In the next stage of our research, we utilized the BERT base model, specifically the uncased version known as bert-base-uncased, to fine-tune with GPT-4 annotated data. BERT (Bidirectional Encoder Representations from Transformers) is renowned in the field of NLP for setting new benchmarks across a variety of tasks. Its architecture incorporates 12 layers of bidirectional transformers, providing a nuanced understanding of contextual information within language.

The uncased model of BERT, pre-trained on an extensive corpus comprising the entire English Wikipedia and BookCorpus, utilizes self-supervised learning with two primary objectives: Masked Language Modeling (MLM) and Next Sentence Prediction (NSP). These pre-training tasks allow BERT to develop a profound bidirectional representation of the English language, crucial for understanding the intricacies of natural language.
BERT's ability to process context effectively, coupled with its relatively compact architecture, makes it particularly suitable for scenarios with limited computational resources or in closed-network environments. In such cases, BERT can capitalize on the rich, context-aware insights extracted from LLMs like GPT-4 to enhance its performance significantly.

In the token classification task, the network we use harnesses the power of BERT's transformer architecture. It begins with an embedding layer that combines word, position, and token type embeddings into a unified representation for each token. This is followed by a series of 12 transformer layers, each applying self-attention that dynamically weights the influence of different parts of the input sentence, refining the token representations based on the context provided by surrounding words.

The contextual representation of each token, h, is computed as a result of successive transformations applied by the encoder layers. Each token embedding is first passed through a series of self-attention heads within each transformer layer, which compute attention scores that reflect the token's relationship with every other token in the sequence. The attention mechanism's output for each token is a weighted sum of the value vectors, which is mathematically represented as:
\begin{equation*}
    Attention(Q, K, V) = \text{softmax}\left(\frac{QK^T}{\sqrt{d_k}}\right)V
\end{equation*}
where Q, K, and V are the query, key, and value matrices derived from the token embeddings, and $d_k$ is the dimensionality of the key vectors. The outputs from the attention heads are then concatenated and passed through a feed-forward neural network, then the token representations are normalized and passed up to the next transformer layer for further refinement. Upon completion of this process through all transformer layers, the final contextualized token representation h is obtained, which encapsulates both the semantic and syntactic information pertinent to each token within its sentence context. The classification layer then takes this representation and applies a linear transformation followed by a softmax function to produce a probability distribution over the possible entity tags:
\begin{equation*}
P(\text{entity tag}|\text{token}) = \text{softmax}(W_c h + b_c)
\end{equation*}
where $W_c$ and $b_c$ are the weights and bias of the classifier layer, and h is the output from the final transformer layer.

\subsubsection*{Experiment Design}
In the second phase of our comparative training experiments, we aimed to assess the effectiveness of using GPT-4 annotated data, termed as 'distilled data', in conjunction with original data to enhance the BERT model's performance on NER tasks.

\textbf{Preparation: }We began by sourcing additional data for annotation from the BBC news dataset \cite{greene2006practical}, chosen for its topical similarity to the CONLL2003 dataset, which is also centered around news content and used to conducting research of document clustering in the original paper. This step was based on the premise that similar data types could bolster the model's training when later combined with the original CONLL data. A pivotal observation from the previous phase indicated that longer sentences provided GPT-4 with the necessary context to produce more accurate annotations. Therefore, we specifically selected longer sentences for GPT-4 to annotate using the Chain of Thought (CoT) prompting method. After the annotation process and removing samples with formatting issues or lacking entities, we compiled a final set of 966 entries, distributed across categories: business, entertainment, politics, sport, and technology.

\textbf{Experimental Setup: }The experiment was structured into several groups, each representing a unique training regimen:
\begin{itemize}
\item Group A utilized 1000 original sentences from the CONLL2003 dataset, training the BERT model for 20 epochs, to serve as a baseline representing training with human-annotated data only.
\item Group B involved training the model for 20 epochs solely on 1000 distilled sentences from the CONLL2003 dataset, providing insight into the utility of LLM-generated annotations.
\item Group C aimed to evaluate the impact of combining distilled data with data from a similar domain, training BERT for 20 epochs on 1000 distilled CONLL sentences augmented with 966 distilled BBC sentences.
\item Group D explored a sequential training approach, initially using 1000 distilled CONLL sentences for 10 epochs, followed by 1000 original CONLL sentences for a subsequent 10 epochs.
\item Group E followed a similar sequential pattern as Group D but started with the combined dataset of 1000 distilled CONLL sentences plus 966 distilled BBC sentences for the first 10 epochs, before proceeding with the 1000 original CONLL sentences for the remaining 10 epochs.
\end{itemize}

For each group, we employed a consistent set of hyperparameters to ensure comparability across results. The training and validation batches were set to sizes of 4 and 2, respectively, with a sequence length capped at 128 tokens. For each group, a fixed learning rate of 1e-05 and learning rate decay strategy with decreasing factor of 0.95 after each epoch are conducted respectively. 

To address variability and ensure the robustness of our findings, each experimental setup was subjected to five iterations of training and testing. The average performance across these iterations was calculated and presented as the final result. The testing was conducted on the CONLL2003 test set, which comprises 3452 data points, allowing us to gauge the model's NER capabilities after training under each experimental condition.

By conducting 50 distinct training cycles across various groupings and learning rate strategies, we sought to thoroughly investigate the potential synergies between distilled and original datasets in improving the BERT model's NER performance.

\subsubsection*{Results and Analysis}
The analysis of the second phase's experimental results (see Table~\ref{tab:p2_table_temp}) reveals several key insights into the performance of different training strategies on the BERT model for NER tasks.

\textbf{Comparison of Pure Distilled Data Groups: }Group B, trained only with distilled CONLL data, showed a micro average F1-score of 0.683 without learning rate decay, which is marginally lower than Group C’s score of 0.713 when trained on both CONLL and BBC distilled data without learning rate decay. This suggests that adding distilled data from a related external source (BBC) can enhance the model's ability to generalize, as evidenced by the improvement in F1-scores.

\textbf{Group C vs. Group A (Pure Original Data): }Group C’s performance, even with the addition of BBC distilled data, did not surpass Group A, which trained purely on original data and achieved a micro average F1-score of 0.850 without learning rate decay. The variance in data sources may contribute to this outcome, emphasizing the importance of consistency.

\textbf{Sequential Training Strategies: }Groups D and E adopted a sequential training strategy, beginning with distilled data and then training on original data. Group D, which mixed CONLL distilled data with original data, showed an improvement in micro average F1-score to 0.859 without learning rate decay. Group E, which included BBC distilled data in the initial training phase, matched this performance, also achieving a micro average F1-score of 0.869 without learning rate decay. These results confirm that a sequential training approach can significantly elevate model performance compared to training solely on original data (Group A).

\textbf{Learning Rate Decay Influence: }The non-learning rate decay setup was generally more favorable for the sequential training groups. For Group E, many scores, e.g. the micro average F1-score and precision and the F1-score of LOC etc, are higher without decay. This pattern indicates that a consistent learning rate might be beneficial when transitioning from distilled to original data during training.

From these observations, it is evident that incorporating distilled data, particularly through a sequential training strategy, can yield significant improvements in NER performance. Moreover, the use of related external datasets as additional distilled data sources has the potential to further improve the model's training efficacy.




\subsection*{Phase Three: Optimizing Training Strategy}
\subsubsection*{Experiment Design}
In Phase Three, we examine various strategies to integrate distilled data and original data throughout the training process. This phase aims to identify the most effective blending technique for enhancing the BERT model's performance on NER tasks. To manage the proportions of data types throughout the training epochs, we introduce a function as follows:
\begin{align*}
f(t) &= w_0 \\
w_1 &= 1 - w_0
\end{align*}
which determines the weight of distilled data used at any given epoch $t$. The weight of original data is then calculated as $w_1$, establishing a flexible framework that accommodates various blending strategies.

\begin{figure}[hbt!]
\centering
\includegraphics[width=0.95\linewidth]{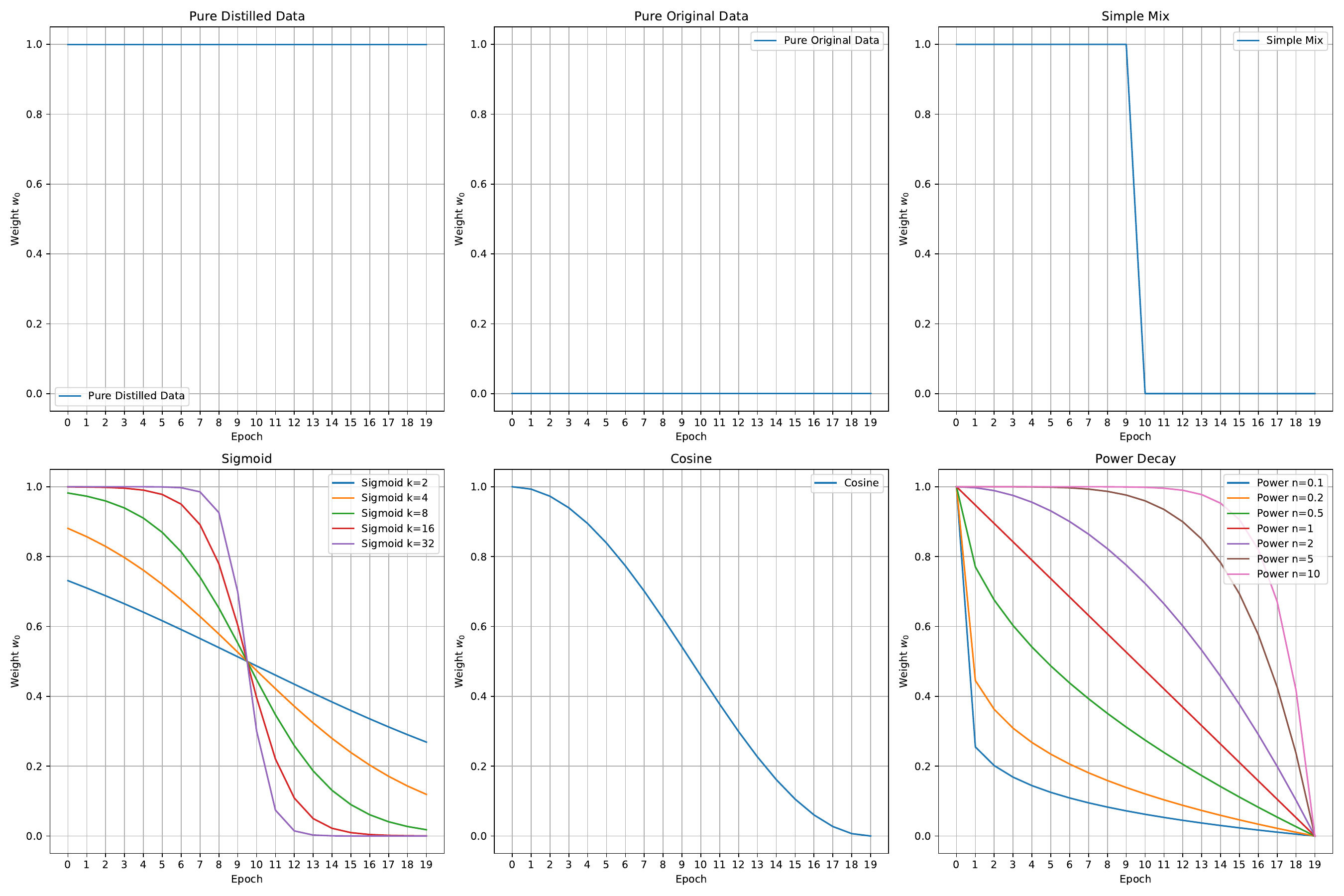}
\caption{Strategies to integrate distilled data and original data}
\label{fig:weight_decay_curves}
\end{figure}

We explore seven distinct blending patterns, each defined by its characteristic function:

\begin{itemize}
\item \textbf{Pure Distilled Data(refer to Group C at phase two): }For all epochs, the distilled data weight, $w_0$, remains at 1, signifying exclusive use of distilled data.
\item \textbf{Pure Original Data(refer to Group A at phase two): }Conversely, $w_0$ stays at 0, indicating the training uses only original data.
\item \textbf{Simple Mix(refer to Group E at phase two): }A direct transition is employed where $w_0$ is 1 for the first 10 epochs and 0 for the second 10 epochs, indicating an initial training with distilled data followed by training with original data. And the number of total epochs is represented with $T$. 
\begin{equation*}
    w_0 =    \begin{cases}    1 & \text{if } t < \frac{T}{2} \\   0 & \text{otherwise}   \end{cases}
\end{equation*}
\item \textbf{Sigmoid Decay: }Here, $w_0$ follows a sigmoid function, smoothly transitioning from 1 to 0 as epochs progress, with a variable k determining the steepness of the curve. This pattern is represented by the equation: 
\begin{align*}
w_0 &= 1 - \frac{1}{1 + e^{-k(x - 0.5)}} \\
x &= \frac{t}{T - 1}
\end{align*}
The experiment utilizes sigmoid functions with different $k$ values to observe the impact of transition steepness on training efficacy. Specifically, $k=\{2,4,8,16,32\}$ in our experiment.
\item \textbf{Cosine Decay: }Utilizing a cosine function, $w_0$ decreases in a sinusoidal pattern, providing a smooth and periodic blending of data. The function is defined as: 
\begin{equation*}
    w_0 = 0.5 \left(1 + \cos\left(\pi \frac{t}{T - 1}\right)\right)
\end{equation*}
\item \textbf{Power Decay: }This method offers a variety of decay rates with the equation: 
\begin{equation*}
    w_0 = 1 - \left(\frac{t}{T - 1}\right)^n
\end{equation*}
where different values of $n$ control the rate at which $w_0$ approaches 0. Specifically, $n=\{0.1,0.2,0.5,1,2,5,10\}$ in our experiment. 
\item \textbf{Full Blend (ALL): }Not depicted in the graph, this approach completely mixes distilled and original data across all epochs, with $w_0 = w_1 = 1$. 
\end{itemize}
Each strategy's parameters are informed by the figure (see Figure~\ref{fig:weight_decay_curves}), visually illustrating the transformation patterns of $w_0$ across epochs for sigmoid, cosine, and power decay functions.

The training details for each pattern remain consistent with the previous phase, using the BERT model and the same hyperparameters, with a distilled dataset of 1966 sentences and an original dataset of 1000 CONLL training sentences. The learning rate decay is also applied as before. 

\subsubsection*{Results and Analysis}
The analysis of the third phase of our experiment, focusing on optimizing training strategies for BERT on NER tasks, reveals some intriguing findings. Here's an in-depth look at the specific results presented in the table(see Figure~\ref{tab:p3_table_temp}):

\textbf{Simple Mix Strategy Analysis: }The 'Simple Mix' approach without learning rate (LR) decay leads to the highest overall performance. Specifically, it achieves a micro-average F1-score of 0.869, a precision of 0.864, and a recall of 0.874. Notably, the LOC category benefits significantly from this strategy, with precision reaching 0.888 and recall at 0.913. However, this strategy does not always dominate across all metrics. For instance, the sigmoid approach with k=32 outperforms it on the macro-average recall with a score of 0.855 and the PER precision with a notable score of 0.957, suggesting that the sigmoid approach may offer competitive advantages in certain scenarios.

\textbf{Learning Rate Decay Impact: }Comparing the results with and without LR decay within the 'Simple Mix' strategy, the non-decay approach generally yields better performance, as evidenced by the higher precision scores across most categories. This suggests that maintaining a constant learning rate might be more beneficial for model stability in certain contexts.

\textbf{Sigmoid Strategy Evaluation: }The sigmoid strategy with larger 'k' values tends to perform better, approaching the 'Simple Mix' results as k increases to 32. The sigmoid curve with k=32 is getting closer to the 'Simple Mix' strategy in terms of performance, suggesting that a more gradual transition between the distilled and original data might provide a more effective learning trajectory.

\textbf{Power Strategy Insights: }For the power strategy, n=0.2 seems to offer the best overall results. This configuration leads to the highest scores in PER F1 (0.956), ORG F1 (0.849), and ORG recall (0.853). This might indicate that a moderately rapid shift towards the original data early in training benefits the model, possibly due to the distinct distributions of the original and distilled data.

\textbf{Cosine Strategy Results: }The cosine strategy provides moderate outcomes. The smooth transition offered by this strategy does not seem to offer any significant advantage over the others, indicating that it may not be the most effective for this application.

\textbf{Comparison with ALL Strategy: }The 'ALL' strategy, which combines both datasets in full for each training epoch, shows the least effective performance. It suggests that simply mixing a larger volume of data from different distributions without a strategic approach can degrade model performance. This highlights the importance of a targeted training strategy over mere data quantity.

From these observations, we can conclude that while the 'Simple Mix' strategy without LR decay stands out for its robust performance, particularly in the LOC and PER categories, alternative strategies like sigmoid (with a higher 'k') and power (with n around 0.2) have their merits. The 'ALL' strategy's underperformance reinforces the need for strategic data blending rather than relying on data quantity alone. Moving forward, these insights will be invaluable for developing more nuanced training strategies that consider the nature and source of training data to maximize performance on NER tasks.


\begin{landscape}
\begin{table}[]
\resizebox{\columnwidth}{!}{%
\begin{tabular}{|cc|cc|cc|cc|cc|cc|}
\hline
\multicolumn{2}{|c|}{} &
  \multicolumn{2}{c|}{\textbf{\begin{tabular}[c]{@{}c@{}}Group B\\ (pure CONLL distilled data)\end{tabular}}} &
  \multicolumn{2}{c|}{\textbf{\begin{tabular}[c]{@{}c@{}}Group C\\ (CONLL + BBC distilled data)\end{tabular}}} &
  \multicolumn{2}{c|}{\textbf{\begin{tabular}[c]{@{}c@{}}Group A\\ (pure original data)\end{tabular}}} &
  \multicolumn{2}{c|}{\textbf{\begin{tabular}[c]{@{}c@{}}Group D\\ (simple mix with CONLL only)\end{tabular}}} &
  \multicolumn{2}{c|}{\textbf{\begin{tabular}[c]{@{}c@{}}Group E\\ (simple mix with CONLL + BBC)\end{tabular}}} \\ \cline{3-12} 
\multicolumn{2}{|c|}{\multirow{-2}{*}{}} &
  \multicolumn{1}{c|}{\textbf{no LR decay}} &
  \textbf{LR decay} &
  \multicolumn{1}{c|}{\textbf{no LR decay}} &
  \textbf{LR decay} &
  \multicolumn{1}{c|}{\textbf{no LR decay}} &
  \textbf{LR decay} &
  \multicolumn{1}{c|}{\textbf{no LR decay}} &
  \textbf{LR decay} &
  \multicolumn{1}{c|}{\textbf{no LR decay}} &
  \textbf{LR decay} \\ \hline
\multicolumn{1}{|c|}{} &
  \textbf{f1-score} &
  \multicolumn{1}{c|}{0.683} &
  0.672 &
  \multicolumn{1}{c|}{0.713} &
  0.691 &
  \multicolumn{1}{c|}{0.850} &
  0.849 &
  \multicolumn{1}{c|}{0.859} &
  0.853 &
  \multicolumn{1}{c|}{{\color[HTML]{FF0000} 0.869}} &
  0.869 \\ \cline{2-12} 
\multicolumn{1}{|c|}{} &
  \textbf{precision} &
  \multicolumn{1}{c|}{0.617} &
  0.597 &
  \multicolumn{1}{c|}{0.652} &
  0.618 &
  \multicolumn{1}{c|}{0.842} &
  0.841 &
  \multicolumn{1}{c|}{0.851} &
  0.846 &
  \multicolumn{1}{c|}{{\color[HTML]{FF0000} 0.864}} &
  0.859 \\ \cline{2-12} 
\multicolumn{1}{|c|}{\multirow{-3}{*}{\textbf{\begin{tabular}[c]{@{}c@{}}micro avg\\ (support: 8598)\end{tabular}}}} &
  \textbf{recall} &
  \multicolumn{1}{c|}{0.766} &
  0.768 &
  \multicolumn{1}{c|}{0.788} &
  0.783 &
  \multicolumn{1}{c|}{0.859} &
  0.856 &
  \multicolumn{1}{c|}{0.868} &
  0.859 &
  \multicolumn{1}{c|}{0.874} &
  {\color[HTML]{FF0000} 0.879} \\ \hline
\multicolumn{1}{|c|}{} &
  \textbf{f1-score} &
  \multicolumn{1}{c|}{0.660} &
  0.655 &
  \multicolumn{1}{c|}{0.684} &
  0.674 &
  \multicolumn{1}{c|}{0.820} &
  0.818 &
  \multicolumn{1}{c|}{0.829} &
  0.820 &
  \multicolumn{1}{c|}{{\color[HTML]{FF0000} 0.840}} &
  0.839 \\ \cline{2-12} 
\multicolumn{1}{|c|}{} &
  \textbf{precision} &
  \multicolumn{1}{c|}{0.639} &
  0.634 &
  \multicolumn{1}{c|}{0.659} &
  0.654 &
  \multicolumn{1}{c|}{0.809} &
  0.807 &
  \multicolumn{1}{c|}{0.818} &
  0.810 &
  \multicolumn{1}{c|}{{\color[HTML]{FF0000} 0.830}} &
  0.825 \\ \cline{2-12} 
\multicolumn{1}{|c|}{\multirow{-3}{*}{\textbf{\begin{tabular}[c]{@{}c@{}}macro avg\\ (support: 8598)\end{tabular}}}} &
  \textbf{recall} &
  \multicolumn{1}{c|}{0.728} &
  0.729 &
  \multicolumn{1}{c|}{0.747} &
  0.748 &
  \multicolumn{1}{c|}{0.834} &
  0.830 &
  \multicolumn{1}{c|}{0.844} &
  0.832 &
  \multicolumn{1}{c|}{0.853} &
  {\color[HTML]{FF0000} 0.855} \\ \hline
\multicolumn{1}{|c|}{} &
  \textbf{f1-score} &
  \multicolumn{1}{c|}{0.735} &
  0.732 &
  \multicolumn{1}{c|}{0.756} &
  0.750 &
  \multicolumn{1}{c|}{0.851} &
  0.850 &
  \multicolumn{1}{c|}{0.861} &
  0.854 &
  \multicolumn{1}{c|}{{\color[HTML]{FF0000} 0.871}} &
  0.871 \\ \cline{2-12} 
\multicolumn{1}{|c|}{} &
  \textbf{precision} &
  \multicolumn{1}{c|}{0.728} &
  0.724 &
  \multicolumn{1}{c|}{0.745} &
  0.744 &
  \multicolumn{1}{c|}{0.845} &
  0.845 &
  \multicolumn{1}{c|}{0.856} &
  0.851 &
  \multicolumn{1}{c|}{{\color[HTML]{FF0000} 0.869}} &
  0.864 \\ \cline{2-12} 
\multicolumn{1}{|c|}{\multirow{-3}{*}{\textbf{\begin{tabular}[c]{@{}c@{}}weighted avg\\ (support: 8598)\end{tabular}}}} &
  \textbf{recall} &
  \multicolumn{1}{c|}{0.766} &
  0.768 &
  \multicolumn{1}{c|}{0.788} &
  0.783 &
  \multicolumn{1}{c|}{0.859} &
  0.856 &
  \multicolumn{1}{c|}{0.868} &
  0.859 &
  \multicolumn{1}{c|}{0.874} &
  {\color[HTML]{FF0000} 0.879} \\ \hline
\multicolumn{1}{|c|}{} &
  \textbf{f1-score} &
  \multicolumn{1}{c|}{0.768} &
  0.776 &
  \multicolumn{1}{c|}{0.807} &
  0.790 &
  \multicolumn{1}{c|}{0.873} &
  0.870 &
  \multicolumn{1}{c|}{0.877} &
  0.875 &
  \multicolumn{1}{c|}{{\color[HTML]{FF0000} 0.897}} &
  0.892 \\ \cline{2-12} 
\multicolumn{1}{|c|}{} &
  \textbf{precision} &
  \multicolumn{1}{c|}{0.711} &
  0.719 &
  \multicolumn{1}{c|}{0.762} &
  0.745 &
  \multicolumn{1}{c|}{0.856} &
  0.851 &
  \multicolumn{1}{c|}{0.862} &
  0.854 &
  \multicolumn{1}{c|}{{\color[HTML]{FF0000} 0.888}} &
  0.872 \\ \cline{2-12} 
\multicolumn{1}{|c|}{\multirow{-3}{*}{\textbf{\begin{tabular}[c]{@{}c@{}}LOC\\ (support: 2132)\end{tabular}}}} &
  \textbf{recall} &
  \multicolumn{1}{c|}{0.836} &
  0.841 &
  \multicolumn{1}{c|}{0.858} &
  0.839 &
  \multicolumn{1}{c|}{0.892} &
  0.889 &
  \multicolumn{1}{c|}{0.895} &
  0.897 &
  \multicolumn{1}{c|}{0.907} &
  {\color[HTML]{FF0000} 0.913} \\ \hline
\multicolumn{1}{|c|}{} &
  \textbf{f1-score} &
  \multicolumn{1}{c|}{0.687} &
  0.688 &
  \multicolumn{1}{c|}{0.725} &
  0.719 &
  \multicolumn{1}{c|}{0.817} &
  0.818 &
  \multicolumn{1}{c|}{0.834} &
  0.827 &
  \multicolumn{1}{c|}{0.844} &
  {\color[HTML]{FF0000} 0.847} \\ \cline{2-12} 
\multicolumn{1}{|c|}{} &
  \textbf{precision} &
  \multicolumn{1}{c|}{0.738} &
  0.739 &
  \multicolumn{1}{c|}{0.752} &
  0.758 &
  \multicolumn{1}{c|}{0.834} &
  0.836 &
  \multicolumn{1}{c|}{0.847} &
  0.849 &
  \multicolumn{1}{c|}{{\color[HTML]{FF0000} 0.868}} &
  0.859 \\ \cline{2-12} 
\multicolumn{1}{|c|}{\multirow{-3}{*}{\textbf{\begin{tabular}[c]{@{}c@{}}ORG\\ (support: 2669)\end{tabular}}}} &
  \textbf{recall} &
  \multicolumn{1}{c|}{0.644} &
  0.644 &
  \multicolumn{1}{c|}{0.701} &
  0.685 &
  \multicolumn{1}{c|}{0.802} &
  0.802 &
  \multicolumn{1}{c|}{0.822} &
  0.806 &
  \multicolumn{1}{c|}{0.822} &
  {\color[HTML]{FF0000} 0.836} \\ \hline
\multicolumn{1}{|c|}{} &
  \textbf{f1-score} &
  \multicolumn{1}{c|}{0.934} &
  0.928 &
  \multicolumn{1}{c|}{0.925} &
  0.932 &
  \multicolumn{1}{c|}{0.942} &
  0.943 &
  \multicolumn{1}{c|}{0.952} &
  0.949 &
  \multicolumn{1}{c|}{{\color[HTML]{FF0000} 0.953}} &
  0.953 \\ \cline{2-12} 
\multicolumn{1}{|c|}{} &
  \textbf{precision} &
  \multicolumn{1}{c|}{0.941} &
  0.927 &
  \multicolumn{1}{c|}{0.930} &
  0.943 &
  \multicolumn{1}{c|}{0.934} &
  0.939 &
  \multicolumn{1}{c|}{0.951} &
  0.947 &
  \multicolumn{1}{c|}{0.952} &
  {\color[HTML]{FF0000} 0.953} \\ \cline{2-12} 
\multicolumn{1}{|c|}{\multirow{-3}{*}{\textbf{\begin{tabular}[c]{@{}c@{}}PER\\ (support: 2768)\end{tabular}}}} &
  \textbf{recall} &
  \multicolumn{1}{c|}{0.927} &
  0.929 &
  \multicolumn{1}{c|}{0.921} &
  0.922 &
  \multicolumn{1}{c|}{0.951} &
  0.947 &
  \multicolumn{1}{c|}{0.953} &
  0.951 &
  \multicolumn{1}{c|}{{\color[HTML]{FF0000} 0.955}} &
  0.954 \\ \hline
\multicolumn{1}{|c|}{} &
  \textbf{f1-score} &
  \multicolumn{1}{c|}{0.251} &
  0.230 &
  \multicolumn{1}{c|}{0.277} &
  0.257 &
  \multicolumn{1}{c|}{0.649} &
  0.640 &
  \multicolumn{1}{c|}{0.654} &
  0.627 &
  \multicolumn{1}{c|}{{\color[HTML]{FF0000} 0.665}} &
  0.664 \\ \cline{2-12} 
\multicolumn{1}{|c|}{} &
  \textbf{precision} &
  \multicolumn{1}{c|}{0.167} &
  0.150 &
  \multicolumn{1}{c|}{0.191} &
  0.169 &
  \multicolumn{1}{c|}{0.613} &
  0.603 &
  \multicolumn{1}{c|}{0.611} &
  0.588 &
  \multicolumn{1}{c|}{0.614} &
  {\color[HTML]{FF0000} 0.618} \\ \cline{2-12} 
\multicolumn{1}{|c|}{\multirow{-3}{*}{\textbf{\begin{tabular}[c]{@{}c@{}}MISC\\ (support: 1029)\end{tabular}}}} &
  \textbf{recall} &
  \multicolumn{1}{c|}{0.504} &
  0.500 &
  \multicolumn{1}{c|}{0.508} &
  0.548 &
  \multicolumn{1}{c|}{0.690} &
  0.684 &
  \multicolumn{1}{c|}{0.706} &
  0.673 &
  \multicolumn{1}{c|}{{\color[HTML]{FF0000} 0.727}} &
  0.718 \\ \hline
\end{tabular}%
}
\caption{The comparison between different groups at phase two}
\label{tab:p2_table_temp}
\end{table}

\begin{table}[]
\resizebox{\columnwidth}{!}{%
\begin{tabular}{|cc|cc|cc|cc|cc|cc|cc|cc|cc|cc|cc|cc|cc|cc|cc|cc|}
\hline
\multicolumn{2}{|c|}{} &
  \multicolumn{2}{c|}{\textbf{simple mix}} &
  \multicolumn{2}{c|}{\textbf{ALL}} &
  \multicolumn{2}{c|}{\textbf{sigmoid(k=2)}} &
  \multicolumn{2}{c|}{\textbf{sigmoid(k=4)}} &
  \multicolumn{2}{c|}{\textbf{sigmoid(k=8)}} &
  \multicolumn{2}{c|}{\textbf{sigmoid(k=16)}} &
  \multicolumn{2}{c|}{\textbf{sigmoid(k=32)}} &
  \multicolumn{2}{c|}{\textbf{power(n=0.1)}} &
  \multicolumn{2}{c|}{\textbf{power(n=0.2)}} &
  \multicolumn{2}{c|}{\textbf{power(n=0.5)}} &
  \multicolumn{2}{c|}{\textbf{power(n=1)}} &
  \multicolumn{2}{c|}{\textbf{power(n=2)}} &
  \multicolumn{2}{c|}{\textbf{power(n=5)}} &
  \multicolumn{2}{c|}{\textbf{power(n=10)}} &
  \multicolumn{2}{c|}{\textbf{cosine}} \\ \cline{3-32} 
\multicolumn{2}{|c|}{\multirow{-2}{*}{}} &
  \multicolumn{1}{c|}{\textbf{no LR decay}} &
  \textbf{LR decay} &
  \multicolumn{1}{c|}{\textbf{no LR decay}} &
  \textbf{LR decay} &
  \multicolumn{1}{c|}{\textbf{no LR decay}} &
  \textbf{LR decay} &
  \multicolumn{1}{c|}{\textbf{no LR decay}} &
  \textbf{LR decay} &
  \multicolumn{1}{c|}{\textbf{no LR decay}} &
  \textbf{LR decay} &
  \multicolumn{1}{c|}{\textbf{no LR decay}} &
  \textbf{LR decay} &
  \multicolumn{1}{c|}{\textbf{no LR decay}} &
  \textbf{LR decay} &
  \multicolumn{1}{c|}{\textbf{no LR decay}} &
  \textbf{LR decay} &
  \multicolumn{1}{c|}{\textbf{no LR decay}} &
  \textbf{LR decay} &
  \multicolumn{1}{c|}{\textbf{no LR decay}} &
  \textbf{LR decay} &
  \multicolumn{1}{c|}{\textbf{no LR decay}} &
  \textbf{LR decay} &
  \multicolumn{1}{c|}{\textbf{no LR decay}} &
  \textbf{LR decay} &
  \multicolumn{1}{c|}{\textbf{no LR decay}} &
  \textbf{LR decay} &
  \multicolumn{1}{c|}{\textbf{no LR decay}} &
  \textbf{LR decay} &
  \multicolumn{1}{c|}{\textbf{no LR decay}} &
  \textbf{LR decay} \\ \hline
\multicolumn{1}{|c|}{} &
  \textbf{f1-score} &
  \multicolumn{1}{c|}{{\color[HTML]{FF0000} 0.869}} &
  0.869 &
  \multicolumn{1}{c|}{0.792} &
  0.800 &
  \multicolumn{1}{c|}{0.811} &
  0.814 &
  \multicolumn{1}{c|}{0.831} &
  0.824 &
  \multicolumn{1}{c|}{0.847} &
  0.846 &
  \multicolumn{1}{c|}{0.861} &
  0.860 &
  \multicolumn{1}{c|}{0.865} &
  0.864 &
  \multicolumn{1}{c|}{0.859} &
  0.856 &
  \multicolumn{1}{c|}{0.861} &
  0.861 &
  \multicolumn{1}{c|}{0.846} &
  0.847 &
  \multicolumn{1}{c|}{0.848} &
  0.849 &
  \multicolumn{1}{c|}{0.840} &
  0.852 &
  \multicolumn{1}{c|}{0.855} &
  0.845 &
  \multicolumn{1}{c|}{0.849} &
  0.849 &
  \multicolumn{1}{c|}{0.853} &
  0.854 \\ \cline{2-32} 
\multicolumn{1}{|c|}{} &
  \textbf{precision} &
  \multicolumn{1}{c|}{{\color[HTML]{FF0000} 0.864}} &
  0.859 &
  \multicolumn{1}{c|}{0.747} &
  0.757 &
  \multicolumn{1}{c|}{0.773} &
  0.782 &
  \multicolumn{1}{c|}{0.811} &
  0.796 &
  \multicolumn{1}{c|}{0.825} &
  0.823 &
  \multicolumn{1}{c|}{0.846} &
  0.845 &
  \multicolumn{1}{c|}{0.851} &
  0.853 &
  \multicolumn{1}{c|}{0.850} &
  0.839 &
  \multicolumn{1}{c|}{0.846} &
  0.846 &
  \multicolumn{1}{c|}{0.826} &
  0.826 &
  \multicolumn{1}{c|}{0.826} &
  0.831 &
  \multicolumn{1}{c|}{0.819} &
  0.834 &
  \multicolumn{1}{c|}{0.841} &
  0.826 &
  \multicolumn{1}{c|}{0.839} &
  0.842 &
  \multicolumn{1}{c|}{0.834} &
  0.836 \\ \cline{2-32} 
\multicolumn{1}{|c|}{\multirow{-3}{*}{\textbf{\begin{tabular}[c]{@{}c@{}}micro avg\\ (support: 8598)\end{tabular}}}} &
  \textbf{recall} &
  \multicolumn{1}{c|}{0.874} &
  {\color[HTML]{FF0000} 0.879} &
  \multicolumn{1}{c|}{0.843} &
  0.848 &
  \multicolumn{1}{c|}{0.854} &
  0.849 &
  \multicolumn{1}{c|}{0.852} &
  0.854 &
  \multicolumn{1}{c|}{0.870} &
  0.869 &
  \multicolumn{1}{c|}{0.877} &
  0.876 &
  \multicolumn{1}{c|}{0.879} &
  0.875 &
  \multicolumn{1}{c|}{0.870} &
  0.874 &
  \multicolumn{1}{c|}{0.876} &
  0.876 &
  \multicolumn{1}{c|}{0.866} &
  0.869 &
  \multicolumn{1}{c|}{0.872} &
  0.867 &
  \multicolumn{1}{c|}{0.862} &
  0.871 &
  \multicolumn{1}{c|}{0.870} &
  0.864 &
  \multicolumn{1}{c|}{0.860} &
  0.857 &
  \multicolumn{1}{c|}{0.873} &
  0.872 \\ \hline
\multicolumn{1}{|c|}{} &
  \textbf{f1-score} &
  \multicolumn{1}{c|}{{\color[HTML]{FF0000} 0.840}} &
  0.839 &
  \multicolumn{1}{c|}{0.755} &
  0.762 &
  \multicolumn{1}{c|}{0.774} &
  0.773 &
  \multicolumn{1}{c|}{0.792} &
  0.786 &
  \multicolumn{1}{c|}{0.810} &
  0.809 &
  \multicolumn{1}{c|}{0.828} &
  0.827 &
  \multicolumn{1}{c|}{0.835} &
  0.833 &
  \multicolumn{1}{c|}{0.827} &
  0.823 &
  \multicolumn{1}{c|}{0.828} &
  0.826 &
  \multicolumn{1}{c|}{0.811} &
  0.811 &
  \multicolumn{1}{c|}{0.813} &
  0.812 &
  \multicolumn{1}{c|}{0.805} &
  0.817 &
  \multicolumn{1}{c|}{0.822} &
  0.808 &
  \multicolumn{1}{c|}{0.816} &
  0.812 &
  \multicolumn{1}{c|}{0.819} &
  0.818 \\ \cline{2-32} 
\multicolumn{1}{|c|}{} &
  \textbf{precision} &
  \multicolumn{1}{c|}{{\color[HTML]{FF0000} 0.830}} &
  0.825 &
  \multicolumn{1}{c|}{0.727} &
  0.734 &
  \multicolumn{1}{c|}{0.745} &
  0.747 &
  \multicolumn{1}{c|}{0.776} &
  0.761 &
  \multicolumn{1}{c|}{0.786} &
  0.786 &
  \multicolumn{1}{c|}{0.810} &
  0.809 &
  \multicolumn{1}{c|}{0.818} &
  0.818 &
  \multicolumn{1}{c|}{0.814} &
  0.804 &
  \multicolumn{1}{c|}{0.809} &
  0.809 &
  \multicolumn{1}{c|}{0.790} &
  0.788 &
  \multicolumn{1}{c|}{0.788} &
  0.793 &
  \multicolumn{1}{c|}{0.782} &
  0.797 &
  \multicolumn{1}{c|}{0.806} &
  0.788 &
  \multicolumn{1}{c|}{0.805} &
  0.802 &
  \multicolumn{1}{c|}{0.798} &
  0.799 \\ \cline{2-32} 
\multicolumn{1}{|c|}{\multirow{-3}{*}{\textbf{\begin{tabular}[c]{@{}c@{}}macro avg\\ (support: 8598)\end{tabular}}}} &
  \textbf{recall} &
  \multicolumn{1}{c|}{0.853} &
  0.855 &
  \multicolumn{1}{c|}{0.813} &
  0.818 &
  \multicolumn{1}{c|}{0.826} &
  0.815 &
  \multicolumn{1}{c|}{0.821} &
  0.826 &
  \multicolumn{1}{c|}{0.840} &
  0.840 &
  \multicolumn{1}{c|}{0.851} &
  0.850 &
  \multicolumn{1}{c|}{{\color[HTML]{FF0000} 0.855}} &
  0.850 &
  \multicolumn{1}{c|}{0.842} &
  0.847 &
  \multicolumn{1}{c|}{0.851} &
  0.847 &
  \multicolumn{1}{c|}{0.842} &
  0.841 &
  \multicolumn{1}{c|}{0.846} &
  0.839 &
  \multicolumn{1}{c|}{0.836} &
  0.844 &
  \multicolumn{1}{c|}{0.843} &
  0.835 &
  \multicolumn{1}{c|}{0.830} &
  0.825 &
  \multicolumn{1}{c|}{0.848} &
  0.844 \\ \hline
\multicolumn{1}{|c|}{} &
  \textbf{f1-score} &
  \multicolumn{1}{c|}{{\color[HTML]{FF0000} 0.871}} &
  0.871 &
  \multicolumn{1}{c|}{0.814} &
  0.821 &
  \multicolumn{1}{c|}{0.827} &
  0.827 &
  \multicolumn{1}{c|}{0.839} &
  0.835 &
  \multicolumn{1}{c|}{0.851} &
  0.852 &
  \multicolumn{1}{c|}{0.864} &
  0.864 &
  \multicolumn{1}{c|}{0.867} &
  0.866 &
  \multicolumn{1}{c|}{0.862} &
  0.860 &
  \multicolumn{1}{c|}{0.864} &
  0.864 &
  \multicolumn{1}{c|}{0.852} &
  0.853 &
  \multicolumn{1}{c|}{0.854} &
  0.855 &
  \multicolumn{1}{c|}{0.845} &
  0.857 &
  \multicolumn{1}{c|}{0.858} &
  0.849 &
  \multicolumn{1}{c|}{0.850} &
  0.851 &
  \multicolumn{1}{c|}{0.858} &
  0.858 \\ \cline{2-32} 
\multicolumn{1}{|c|}{} &
  \textbf{precision} &
  \multicolumn{1}{c|}{{\color[HTML]{FF0000} 0.869}} &
  0.864 &
  \multicolumn{1}{c|}{0.801} &
  0.808 &
  \multicolumn{1}{c|}{0.813} &
  0.813 &
  \multicolumn{1}{c|}{0.833} &
  0.824 &
  \multicolumn{1}{c|}{0.836} &
  0.838 &
  \multicolumn{1}{c|}{0.854} &
  0.855 &
  \multicolumn{1}{c|}{0.858} &
  0.858 &
  \multicolumn{1}{c|}{0.855} &
  0.848 &
  \multicolumn{1}{c|}{0.855} &
  0.854 &
  \multicolumn{1}{c|}{0.843} &
  0.841 &
  \multicolumn{1}{c|}{0.841} &
  0.846 &
  \multicolumn{1}{c|}{0.833} &
  0.845 &
  \multicolumn{1}{c|}{0.848} &
  0.838 &
  \multicolumn{1}{c|}{0.843} &
  0.847 &
  \multicolumn{1}{c|}{0.847} &
  0.848 \\ \cline{2-32} 
\multicolumn{1}{|c|}{\multirow{-3}{*}{\textbf{\begin{tabular}[c]{@{}c@{}}weighted avg\\ (support: 8598)\end{tabular}}}} &
  \textbf{recall} &
  \multicolumn{1}{c|}{0.874} &
  {\color[HTML]{FF0000} 0.879} &
  \multicolumn{1}{c|}{0.843} &
  0.848 &
  \multicolumn{1}{c|}{0.854} &
  0.849 &
  \multicolumn{1}{c|}{0.852} &
  0.854 &
  \multicolumn{1}{c|}{0.870} &
  0.869 &
  \multicolumn{1}{c|}{0.877} &
  0.876 &
  \multicolumn{1}{c|}{0.879} &
  0.875 &
  \multicolumn{1}{c|}{0.870} &
  0.874 &
  \multicolumn{1}{c|}{0.876} &
  0.876 &
  \multicolumn{1}{c|}{0.866} &
  0.869 &
  \multicolumn{1}{c|}{0.872} &
  0.867 &
  \multicolumn{1}{c|}{0.862} &
  0.871 &
  \multicolumn{1}{c|}{0.870} &
  0.864 &
  \multicolumn{1}{c|}{0.860} &
  0.857 &
  \multicolumn{1}{c|}{0.873} &
  0.872 \\ \hline
\multicolumn{1}{|c|}{} &
  \textbf{f1-score} &
  \multicolumn{1}{c|}{{\color[HTML]{FF0000} 0.897}} &
  0.892 &
  \multicolumn{1}{c|}{0.855} &
  0.858 &
  \multicolumn{1}{c|}{0.864} &
  0.864 &
  \multicolumn{1}{c|}{0.868} &
  0.869 &
  \multicolumn{1}{c|}{0.881} &
  0.884 &
  \multicolumn{1}{c|}{0.895} &
  0.890 &
  \multicolumn{1}{c|}{0.890} &
  0.893 &
  \multicolumn{1}{c|}{0.890} &
  0.889 &
  \multicolumn{1}{c|}{0.888} &
  0.895 &
  \multicolumn{1}{c|}{0.883} &
  0.884 &
  \multicolumn{1}{c|}{0.883} &
  0.884 &
  \multicolumn{1}{c|}{0.871} &
  0.883 &
  \multicolumn{1}{c|}{0.888} &
  0.874 &
  \multicolumn{1}{c|}{0.882} &
  0.876 &
  \multicolumn{1}{c|}{0.886} &
  0.891 \\ \cline{2-32} 
\multicolumn{1}{|c|}{} &
  \textbf{precision} &
  \multicolumn{1}{c|}{{\color[HTML]{FF0000} 0.888}} &
  0.872 &
  \multicolumn{1}{c|}{0.828} &
  0.827 &
  \multicolumn{1}{c|}{0.833} &
  0.840 &
  \multicolumn{1}{c|}{0.864} &
  0.842 &
  \multicolumn{1}{c|}{0.854} &
  0.869 &
  \multicolumn{1}{c|}{0.878} &
  0.877 &
  \multicolumn{1}{c|}{0.871} &
  0.875 &
  \multicolumn{1}{c|}{0.880} &
  0.879 &
  \multicolumn{1}{c|}{0.871} &
  0.884 &
  \multicolumn{1}{c|}{0.859} &
  0.870 &
  \multicolumn{1}{c|}{0.859} &
  0.867 &
  \multicolumn{1}{c|}{0.843} &
  0.868 &
  \multicolumn{1}{c|}{0.883} &
  0.853 &
  \multicolumn{1}{c|}{0.868} &
  0.862 &
  \multicolumn{1}{c|}{0.871} &
  0.878 \\ \cline{2-32} 
\multicolumn{1}{|c|}{\multirow{-3}{*}{\textbf{\begin{tabular}[c]{@{}c@{}}LOC\\ (support: 2132)\end{tabular}}}} &
  \textbf{recall} &
  \multicolumn{1}{c|}{0.907} &
  {\color[HTML]{FF0000} 0.913} &
  \multicolumn{1}{c|}{0.884} &
  0.893 &
  \multicolumn{1}{c|}{0.898} &
  0.889 &
  \multicolumn{1}{c|}{0.874} &
  0.897 &
  \multicolumn{1}{c|}{0.909} &
  0.900 &
  \multicolumn{1}{c|}{{\color[HTML]{FF0000} 0.913}} &
  0.903 &
  \multicolumn{1}{c|}{0.910} &
  0.912 &
  \multicolumn{1}{c|}{0.900} &
  0.900 &
  \multicolumn{1}{c|}{0.905} &
  0.906 &
  \multicolumn{1}{c|}{0.909} &
  0.898 &
  \multicolumn{1}{c|}{0.908} &
  0.902 &
  \multicolumn{1}{c|}{0.903} &
  0.900 &
  \multicolumn{1}{c|}{0.894} &
  0.896 &
  \multicolumn{1}{c|}{0.898} &
  0.891 &
  \multicolumn{1}{c|}{0.903} &
  0.904 \\ \hline
\multicolumn{1}{|c|}{} &
  \textbf{f1-score} &
  \multicolumn{1}{c|}{0.844} &
  0.847 &
  \multicolumn{1}{c|}{0.797} &
  0.804 &
  \multicolumn{1}{c|}{0.810} &
  0.808 &
  \multicolumn{1}{c|}{0.823} &
  0.815 &
  \multicolumn{1}{c|}{0.831} &
  0.832 &
  \multicolumn{1}{c|}{0.842} &
  0.846 &
  \multicolumn{1}{c|}{0.841} &
  0.841 &
  \multicolumn{1}{c|}{0.839} &
  0.836 &
  \multicolumn{1}{c|}{0.843} &
  {\color[HTML]{FF0000} 0.849} &
  \multicolumn{1}{c|}{0.828} &
  0.837 &
  \multicolumn{1}{c|}{0.838} &
  0.837 &
  \multicolumn{1}{c|}{0.823} &
  0.837 &
  \multicolumn{1}{c|}{0.835} &
  0.832 &
  \multicolumn{1}{c|}{0.822} &
  0.830 &
  \multicolumn{1}{c|}{0.837} &
  0.837 \\ \cline{2-32} 
\multicolumn{1}{|c|}{} &
  \textbf{precision} &
  \multicolumn{1}{c|}{{\color[HTML]{FF0000} 0.868}} &
  0.859 &
  \multicolumn{1}{c|}{0.822} &
  0.832 &
  \multicolumn{1}{c|}{0.834} &
  0.813 &
  \multicolumn{1}{c|}{0.835} &
  0.840 &
  \multicolumn{1}{c|}{0.834} &
  0.824 &
  \multicolumn{1}{c|}{0.854} &
  0.854 &
  \multicolumn{1}{c|}{0.845} &
  0.852 &
  \multicolumn{1}{c|}{0.848} &
  0.832 &
  \multicolumn{1}{c|}{0.851} &
  0.846 &
  \multicolumn{1}{c|}{0.848} &
  0.846 &
  \multicolumn{1}{c|}{0.853} &
  0.850 &
  \multicolumn{1}{c|}{0.831} &
  0.836 &
  \multicolumn{1}{c|}{0.846} &
  0.843 &
  \multicolumn{1}{c|}{0.833} &
  0.843 &
  \multicolumn{1}{c|}{0.843} &
  0.844 \\ \cline{2-32} 
\multicolumn{1}{|c|}{\multirow{-3}{*}{\textbf{\begin{tabular}[c]{@{}c@{}}ORG\\ (support: 2669)\end{tabular}}}} &
  \textbf{recall} &
  \multicolumn{1}{c|}{0.822} &
  0.836 &
  \multicolumn{1}{c|}{0.776} &
  0.778 &
  \multicolumn{1}{c|}{0.789} &
  0.803 &
  \multicolumn{1}{c|}{0.813} &
  0.792 &
  \multicolumn{1}{c|}{0.829} &
  0.842 &
  \multicolumn{1}{c|}{0.830} &
  0.839 &
  \multicolumn{1}{c|}{0.839} &
  0.831 &
  \multicolumn{1}{c|}{0.831} &
  0.841 &
  \multicolumn{1}{c|}{0.836} &
  {\color[HTML]{FF0000} 0.853} &
  \multicolumn{1}{c|}{0.810} &
  0.828 &
  \multicolumn{1}{c|}{0.825} &
  0.825 &
  \multicolumn{1}{c|}{0.816} &
  0.838 &
  \multicolumn{1}{c|}{0.825} &
  0.822 &
  \multicolumn{1}{c|}{0.813} &
  0.817 &
  \multicolumn{1}{c|}{0.832} &
  0.830 \\ \hline
\multicolumn{1}{|c|}{} &
  \textbf{f1-score} &
  \multicolumn{1}{c|}{0.953} &
  0.953 &
  \multicolumn{1}{c|}{0.946} &
  0.950 &
  \multicolumn{1}{c|}{0.947} &
  0.948 &
  \multicolumn{1}{c|}{0.949} &
  0.948 &
  \multicolumn{1}{c|}{0.950} &
  0.950 &
  \multicolumn{1}{c|}{0.950} &
  0.952 &
  \multicolumn{1}{c|}{0.954} &
  0.949 &
  \multicolumn{1}{c|}{0.947} &
  0.950 &
  \multicolumn{1}{c|}{{\color[HTML]{FF0000} 0.956}} &
  0.950 &
  \multicolumn{1}{c|}{0.952} &
  0.949 &
  \multicolumn{1}{c|}{0.951} &
  0.953 &
  \multicolumn{1}{c|}{0.945} &
  0.951 &
  \multicolumn{1}{c|}{0.943} &
  0.946 &
  \multicolumn{1}{c|}{0.939} &
  0.948 &
  \multicolumn{1}{c|}{0.952} &
  0.952 \\ \cline{2-32} 
\multicolumn{1}{|c|}{} &
  \textbf{precision} &
  \multicolumn{1}{c|}{0.952} &
  0.953 &
  \multicolumn{1}{c|}{0.942} &
  0.947 &
  \multicolumn{1}{c|}{0.941} &
  0.950 &
  \multicolumn{1}{c|}{0.948} &
  0.945 &
  \multicolumn{1}{c|}{0.945} &
  0.956 &
  \multicolumn{1}{c|}{0.943} &
  0.949 &
  \multicolumn{1}{c|}{{\color[HTML]{FF0000} 0.957}} &
  0.946 &
  \multicolumn{1}{c|}{0.942} &
  0.948 &
  \multicolumn{1}{c|}{0.956} &
  0.950 &
  \multicolumn{1}{c|}{0.953} &
  0.942 &
  \multicolumn{1}{c|}{0.944} &
  0.954 &
  \multicolumn{1}{c|}{0.946} &
  0.951 &
  \multicolumn{1}{c|}{0.929} &
  0.940 &
  \multicolumn{1}{c|}{0.928} &
  0.949 &
  \multicolumn{1}{c|}{0.950} &
  0.949 \\ \cline{2-32} 
\multicolumn{1}{|c|}{\multirow{-3}{*}{\textbf{\begin{tabular}[c]{@{}c@{}}PER\\ (support: 2768)\end{tabular}}}} &
  \textbf{recall} &
  \multicolumn{1}{c|}{0.955} &
  0.954 &
  \multicolumn{1}{c|}{0.951} &
  0.954 &
  \multicolumn{1}{c|}{0.953} &
  0.946 &
  \multicolumn{1}{c|}{0.951} &
  0.950 &
  \multicolumn{1}{c|}{0.956} &
  0.945 &
  \multicolumn{1}{c|}{0.958} &
  0.954 &
  \multicolumn{1}{c|}{0.952} &
  0.952 &
  \multicolumn{1}{c|}{0.952} &
  0.952 &
  \multicolumn{1}{c|}{0.955} &
  0.950 &
  \multicolumn{1}{c|}{0.951} &
  0.955 &
  \multicolumn{1}{c|}{0.957} &
  0.951 &
  \multicolumn{1}{c|}{0.944} &
  0.951 &
  \multicolumn{1}{c|}{{\color[HTML]{FF0000} 0.958}} &
  0.953 &
  \multicolumn{1}{c|}{0.952} &
  0.948 &
  \multicolumn{1}{c|}{0.954} &
  0.955 \\ \hline
\multicolumn{1}{|c|}{} &
  \textbf{f1-score} &
  \multicolumn{1}{c|}{{\color[HTML]{FF0000} 0.665}} &
  0.664 &
  \multicolumn{1}{c|}{0.421} &
  0.437 &
  \multicolumn{1}{c|}{0.475} &
  0.474 &
  \multicolumn{1}{c|}{0.527} &
  0.513 &
  \multicolumn{1}{c|}{0.577} &
  0.570 &
  \multicolumn{1}{c|}{0.626} &
  0.620 &
  \multicolumn{1}{c|}{0.654} &
  0.650 &
  \multicolumn{1}{c|}{0.631} &
  0.618 &
  \multicolumn{1}{c|}{0.625} &
  0.611 &
  \multicolumn{1}{c|}{0.581} &
  0.573 &
  \multicolumn{1}{c|}{0.578} &
  0.575 &
  \multicolumn{1}{c|}{0.582} &
  0.597 &
  \multicolumn{1}{c|}{0.623} &
  0.581 &
  \multicolumn{1}{c|}{0.620} &
  0.595 &
  \multicolumn{1}{c|}{0.601} &
  0.593 \\ \cline{2-32} 
\multicolumn{1}{|c|}{} &
  \textbf{precision} &
  \multicolumn{1}{c|}{0.614} &
  {\color[HTML]{FF0000} 0.618} &
  \multicolumn{1}{c|}{0.315} &
  0.331 &
  \multicolumn{1}{c|}{0.372} &
  0.383 &
  \multicolumn{1}{c|}{0.457} &
  0.417 &
  \multicolumn{1}{c|}{0.508} &
  0.496 &
  \multicolumn{1}{c|}{0.564} &
  0.555 &
  \multicolumn{1}{c|}{0.599} &
  0.602 &
  \multicolumn{1}{c|}{0.586} &
  0.557 &
  \multicolumn{1}{c|}{0.559} &
  0.555 &
  \multicolumn{1}{c|}{0.499} &
  0.494 &
  \multicolumn{1}{c|}{0.497} &
  0.499 &
  \multicolumn{1}{c|}{0.509} &
  0.531 &
  \multicolumn{1}{c|}{0.567} &
  0.517 &
  \multicolumn{1}{c|}{0.590} &
  0.555 &
  \multicolumn{1}{c|}{0.527} &
  0.523 \\ \cline{2-32} 
\multicolumn{1}{|c|}{\multirow{-3}{*}{\textbf{\begin{tabular}[c]{@{}c@{}}MISC\\ (support: 1029)\end{tabular}}}} &
  \textbf{recall} &
  \multicolumn{1}{c|}{{\color[HTML]{FF0000} 0.727}} &
  0.718 &
  \multicolumn{1}{c|}{0.640} &
  0.648 &
  \multicolumn{1}{c|}{0.666} &
  0.624 &
  \multicolumn{1}{c|}{0.645} &
  0.667 &
  \multicolumn{1}{c|}{0.668} &
  0.674 &
  \multicolumn{1}{c|}{0.703} &
  0.705 &
  \multicolumn{1}{c|}{0.721} &
  0.706 &
  \multicolumn{1}{c|}{0.684} &
  0.697 &
  \multicolumn{1}{c|}{0.709} &
  0.679 &
  \multicolumn{1}{c|}{0.698} &
  0.683 &
  \multicolumn{1}{c|}{0.693} &
  0.679 &
  \multicolumn{1}{c|}{0.680} &
  0.686 &
  \multicolumn{1}{c|}{0.694} &
  0.668 &
  \multicolumn{1}{c|}{0.659} &
  0.643 &
  \multicolumn{1}{c|}{0.702} &
  0.686 \\ \hline
\end{tabular}%
}
\caption{Comparison of performance between strategies}
\label{tab:p3_table_temp}
\end{table}
\end{landscape}

\section*{Discussion}
In the discussion section of our paper, we delve into the merits and limitations of our proposed solution, as well as outline avenues for future work.

\subsection*{Advantages of the Proposed Solution}
Our approach reduces the need for extensive human annotation, thereby increasing efficiency. The results demonstrate that models trained on a mix of GPT-4 annotated data and human annotations outperform those trained on human annotations alone. This method is particularly advantageous for cold-start scenarios, where using GPT-4 for pre-annotation or training models on pre-annotated data provides them with a foundational capability, aiding annotators in quickly ramping up the annotation process.

The hybrid training of small models with a mix of distilled data and human annotations presents higher adaptability for scenarios with limited resources or closed network environments. This adaptability is critical for applications where large language models cannot be directly accessed. Our experiments also reveal that incorporating distilled data from external sources related to the domain of interest (e.g., BBC news data, in addition to CONLL2003) can lead to performance improvements. This supports the notion that leveraging similar data types for distillation is beneficial, as reflected in our NER task results.

The Chain of Thought (CoT) methodology enhances the interpretability of the annotation process, allowing for easier validation and corrections of the results. The detailed reasoning provided by CoT prompts not only improves the inference performance but also serves an educational purpose, guiding novice annotators in the tagging process.

The use of structured prompts outputting dictionaries, such as {{'LOC': ['Houston'], 'PER': ['Orlando Miller', 'Todd Stottlemyre'], 'ORG': ['Houston Astros', 'St. Louis Cardinals', 'NL Central division']}}, enables the efficient representation of all entity types in a single request. This format can easily handle non-continuous entities and nested entities, and it facilitates automated processing through conversion to dictionary objects. Additionally, the reasoning process uncovers certain annotation issues within the CONLL dataset, where GPT's classifications sometimes appear more logical than the original labels.

\subsection*{Limitations}
The output from LLMs like GPT-4 can still be unstable, occasionally deviating from the desired format or presenting entities that do not match the source text exactly, such as inconsistencies in capitalization or content "corrections" based on the model's interpretation. This necessitates manual review and correction.

The hallucination issue in GPT models persists, with a tendency to over-identify entities, which can reduce the accuracy of the MISC category.

One potential limitation is the dependency on the quality and the relevance of the few-shot examples provided to GPT-4. If the examples are not closely aligned with the target domain or context, the quality of the distilled data may suffer.

While the 'Simple Mix' strategy yielded the most effective results in our experiments, the underlying mechanisms contributing to its success are not fully understood. Future research will delve deeper into this phenomenon to uncover the reasons behind its efficacy and explore whether this pattern holds across different LLMs, datasets, and NLP tasks.

\subsection*{Future Work}
To address the first limitation, future research could explore more advanced LLMs or optimize the phrasing of prompts. Additionally, parameterized prompting could be considered for LLMs deployable in a localized setting.

For the second limitation, we could refine our approach by selecting few-shot examples based on embedding similarity, dynamically choosing examples that are more closely related to the entity types in the target annotation task, thereby improving accuracy.

Implementing a tiered selection process for samples annotated by LLMs might be beneficial. High-quality LLM annotations could be identified and used for further pre-training of smaller models, enhancing their performance.

Integrating active learning strategies where LLMs focus on annotating more challenging samples could boost the efficiency of the smaller model's learning process.

Ongoing optimization of training strategies, exploring new combinations of LLMs and smaller models, and extending the approach to a wider array of NLP tasks are all promising directions for future research.


\section*{Conclusion}
In conclusion, this research substantiates the transformative potential of integrating Large Language Models (LLMs) like GPT-4 into traditional NLP tasks, particularly Named Entity Recognition (NER). By harnessing the Chain of Thought (CoT) prompting, we have effectively distilled LLM knowledge to enhance the training of more compact models such as BERT.

Our investigation has solidified the efficacy of a mixed-training strategy that synergizes LLM-generated annotations with human annotations. This combined approach surpasses the performance of models trained solely on human-annotated data, leading to improvements in efficiency, cost-effectiveness, and model accuracy—goals we set out to achieve by leveraging LLM strengths and mitigating their limitations.

Our method has demonstrated particular value in resource-limited and closed-network settings, providing a versatile solution where the deployment of large-scale LLMs is not viable. The adaptability and practicality of smaller models are thus accentuated, offering a viable pathway to harnessing the power of LLMs in a broader range of contexts.

Reflecting on the challenges encountered, such as LLM variability and hallucination tendencies, we see clear directions for future work. Our research will extend to enhance prompt sophistication, investigate parameterized prompting, and refine the selection of LLM annotations to further elevate NER model quality. We will also delve deeper into optimizing training strategies, as the 'Simple Mix' approach showed the most promise in our studies. Expanding upon these findings, our future endeavors will aim to refine data blending techniques and explore their applicability across a spectrum of NLP tasks, with the goal of continuously pushing the boundaries of what's achievable in the field of natural language processing.




\bibliographystyle{unsrt}  
\bibliography{references}

\end{document}